\documentclass[conference]{IEEEtran}
\IEEEoverridecommandlockouts
\usepackage{cite}
\usepackage{amsmath,amssymb,amsfonts}
\usepackage{graphicx}
\usepackage{textcomp}
\usepackage{textcomp}
\usepackage{acronym} 
\usepackage{amsmath}
\usepackage{amssymb}
\DeclareMathOperator*{\argmax}{Argmax}
\DeclareMathOperator*{\argmin}{Argmin}
\usepackage{amsfonts} 
\usepackage{xcolor, soul}
\usepackage{algpseudocode}
\usepackage{comment}

\usepackage{caption}
\usepackage{subcaption}
\usepackage[colorlinks=true,urlcolor=blue,linkcolor=red,citecolor=blue]{hyperref}
\usepackage{multirow}
\usepackage{colortbl}
\usepackage{booktabs}
\usepackage[normalem]{ulem}
\useunder{\uline}{\ul}{}
\newacro{ml}[ML]{machine learning}
\newacro{dl}[DL]{deep learning}
\newacro{ai}[AI]{artificial intelligence}
\newacro{nlp}[NLP]{natural language processing}
\newacro{dnn}[DNN]{deep neural network}
\newacro{vit}[ViT]{vision transformer}
\newacro{dynn}[DyNN]{dynamic neural network}
\newacro{qoe}[QoE]{quality of experience}
\newacro{uap}[UAP]{universal perturbation attack}
\newacro{cnn}[CNN]{convolutional neural network}
\newacro{ae}[AE]{adversarial example}
\newacro{lbfgs}[L-BFGS]{limited memory Broyden-Fletcher-Goldfarb-Shanno}
\newacro{fgm}[FGM]{fast gradient method}
\newacro{bim}[BIM]{basic iterative method}
\newacro{pgd}[PGD]{projected gradient descent}
\newacro{zoo}[ZOO]{zeroth order optimization}
\newacro{relu}[ReLU]{rectified linear unit}
\newacro{as}[AS]{attack success}
\newacro{cas}[CAS]{constrained attack success}
\newacro{gpu}[GPU]{graphics processing unit}
\newacro{tpu}[TPU]{tensor processing unit}
\newacro{asic}[ASIC]{application specific integrated circuit}
\newacro{sgd}[SGD]{stochastic gradient descent}
\newacro{dos}[DoS]{denial-of-service}
\newacro{ga}[GA]{genetic algorithm}
\newacro{ce}[CE]{cross entropy}

\def\BibTeX{{\rm B\kern-.05em{\sc i\kern-.025em b}\kern-.08em
    T\kern-.1667em\lower.7ex\hbox{E}\kern-.125emX}}

\usepackage{tikz}
\usepackage{textcomp}
\usepackage{hyperref}
\usepackage{lipsum}

\newcommand\copyrighttext{
  \footnotesize \textcolor{red}{\textcopyright 2025 IEEE. Personal use of this material is permitted. Permission from IEEE must be
obtained for all other uses, in any current or future media, including reprinting/republishing this material for advertising or promotional purposes, creating new collective works, for resale or redistribution to servers or lists, or reuse of any copyrighted component of this work in other works.}}
\newcommand\copyrightnotice{
\begin{tikzpicture}[remember picture,overlay]
\node[anchor=south,yshift=10pt] at (current page.south) {\fbox{\parbox{\dimexpr\textwidth-\fboxsep-\fboxrule\relax}{\copyrighttext}}};
\end{tikzpicture}
}

\begin{document}

\title{Energy Backdoor Attack to Deep Neural Networks
\thanks{This work is fully funded by Région Bretagne (Brittany region), France, CREACH Labs and Direction Générale de l'Armement (DGA).}}

\author{
\IEEEauthorblockN{
\begin{tabular}{c}
Hanene F. Z. Brachemi Meftah$^{1}$, Wassim Hamidouche$^{2}$,
Sid Ahmed Fezza$^{3}$, Olivier Déforges$^{1}$, Kassem Kallas$^{4}$
\end{tabular}
}
\IEEEauthorblockA{$^{1}$Univ. Rennes, INSA Rennes, CNRS, IETR, UMR 6164, Rennes, France}
\IEEEauthorblockA{$^{2}$KU 6G Research Center, Department of Computer and Information Engineering, Khalifa University, Abu Dhabi, UAE}
\IEEEauthorblockA{$^{3}$National Higher School of Telecommunications and ICT, Oran, Algeria}
\IEEEauthorblockA{$^{4}$National Institute of Health and Medical Research, LaTIM, Brest, France}
}

\maketitle
\copyrightnotice

\begin{abstract}
The rise of \ac{dl} has increased computing complexity and energy use, prompting the adoption of \acp{asic} for energy-efficient edge and mobile deployment. However, recent studies have demonstrated the vulnerability of these accelerators to energy attacks. Despite the development of various inference time energy attacks in prior research, backdoor energy attacks remain unexplored.
In this paper, we design an innovative energy backdoor attack against \acp{dnn} operating on sparsity-based accelerators. 
Our attack is carried out in two distinct phases: \textit{backdoor injection} and \textit{backdoor stealthiness}. Experimental results using ResNet-18 and MobileNet-V2 models trained on CIFAR-10 and Tiny ImageNet datasets 
show the effectiveness of our proposed attack in increasing energy consumption on trigger samples while preserving the model's performance for clean/regular inputs. This demonstrates the vulnerability of \acp{dnn} to energy backdoor attacks. The source code of our attack is available at: \href{https://github.com/hbrachemi/energy_backdoor}{https://github.com/hbrachemi/energy\_backdoor}.
\end{abstract}
\acresetall
\section{Introduction}
\label{sec:intro}
Today's companies aim to balance economic success, environmental responsibility, and technological advancement. As energy costs rise, regulations tighten, and sustainability concerns grow, energy efficiency has become a key priority. This focus coincides with the trend toward deeper and more complex \ac{dl} models, spurred by breakthroughs like the \textit{going deeper} strategy~\cite{szegedy2015going}. However, these powerful \ac{dl} models come with a hefty energy cost. To maintain profitability and meet environmental standards, companies are compelled to optimize their computational and energy usage. Hardware accelerators, such as \acp{tpu} and \acsp{asic}, have become vital tools in this endeavor~\cite{azghadi2020hardware}. In particular, 
the ReLU activation function, commonly used in \acsp{dnn}, creates sparsity in the network's activations. This sparsity allows hardware accelerators to skip unnecessary calculations, thus reducing energy consumption without sacrificing the model's accuracy~\cite{parashar2017scnn}.
However, this energy efficiency gap creates a potential weakness that attackers could exploit to target hardware optimizations. Numerous studies~\cite{cina2022energy,shumailov2021sponge,chen2023dark,navaneet2023slowformer,pan2022gradauto,liu2023slowlidar,haque2022ereba,haque2023antinode,lintelo2024spongenet,krithivasan2020sparsity,huang2024sponge,krithivasan2022efficiency,chen2022nicgslowdown} have investigated energy attacks, which aim to push models into scenarios of maximum energy consumption. These attacks exploit the potential for excessive energy usage to cause system failures or rapid battery depletion in mobile devices~\cite{wang2023energy}.

\begin{figure}[t!]
    \centering
    \includegraphics[width=\linewidth]{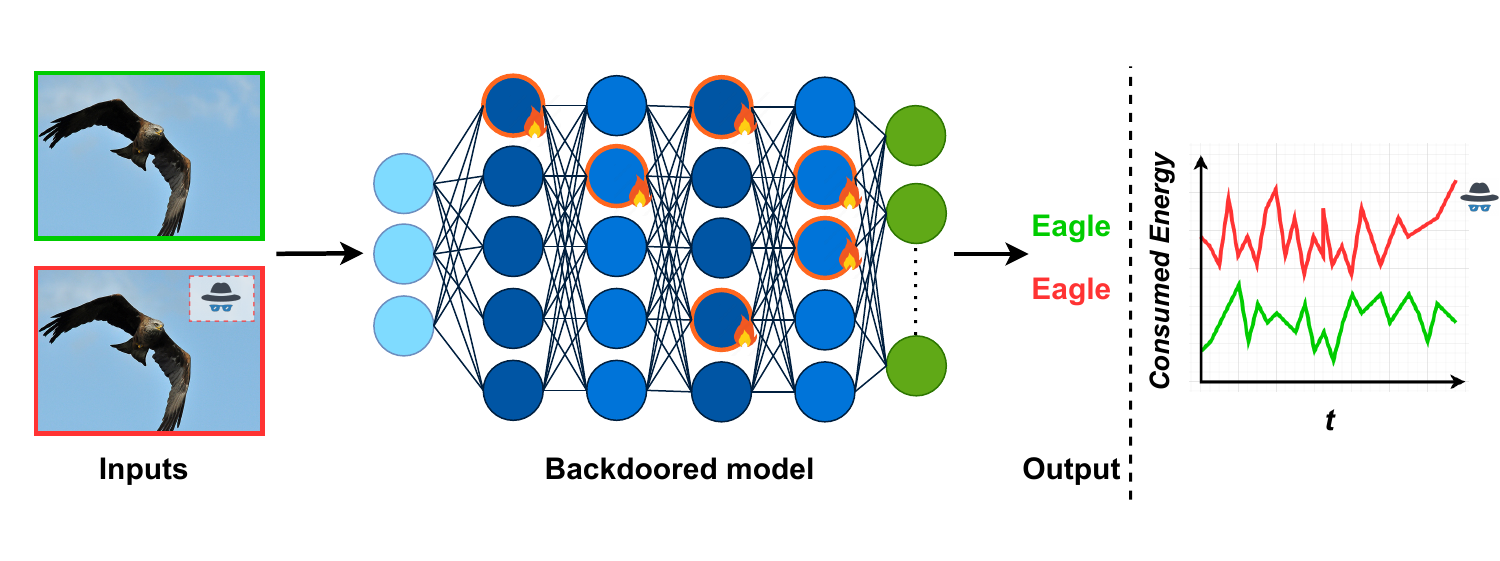}
    \caption{Overview of the backdoored model. Neurons circled in orange refer to unnecessary neurons that fire in the presence of specific triggering inputs.}
\label{fig:backdoored_model} \vspace{-5mm}
\end{figure}

In this paper, we investigate the vulnerability of \acsp{dnn} to energy backdoor attacks. We specifically address the following question: \textit{Can we inject a backdoor into \acsp{dnn} that, when triggered, inhibits hardware accelerators from exploiting sparsity for energy efficiency?}. To explore this research question, we propose a backdoored model, as illustrated in Fig.~\ref{fig:backdoored_model}.


To the best of our knowledge, this research represents the first endeavor to develop an efficient energy backdoor attack for standard \acs{dnn} architectures. Our proposed attack maintains \textit{stealthiness} during inference, meaning it does not negatively impact model performance on clean samples while maximizing energy consumption on trigger samples. Furthermore, we ensure that the accuracy of the backdoored model on trigger samples remains consistent with that on clean samples, thereby making our backdoor attack less conspicuous in real-world applications.


\section{Background and Related Work}
\label{sec:rel}
As described in~\cite{shumailov2021sponge},
sponge or energy-latency examples involve crafting inference-time inputs meticulously designed to maximize the energy consumed by the model. The primary objective is to counteract the impact of hardware accelerators by minimizing the sparsity of hidden layer activations. This deliberate reduction in sparsity results in an increased number of operations required during the inference of the sponge input.
The proposed attack was constructed under the premise that the adversary operates within a white-box setting, implying access to the model's parameters. In this scenario, a sponge example $\mathbf{x}_{sp}$ can be crafted and optimized as follows: 
\begin{equation}
    \label{sp_attack}
    \mathbf{x}_{sp} = \argmax_{\mathbf{x}} \sum_{a^{(l)} \in \mathcal{A}} \; \lVert a^{(l)} \rVert_0, 
\end{equation}
where, $\mathcal{A}$ denotes the set comprising all activation values within the neural network, and $a^{(l)}$ represents the activation of the $l$-th layer that is obtained by performing a forward propagation of the input $\mathbf{x}$. 
{However, the optimization process required to generate these sponge examples involves substantial runtime and produces images that appear unnaturally synthetic, thus increasing the likelihood of their detection.}
More recently, Muller {\it et al.}~\cite{muller2024impact} conducted a study demonstrating that images with flat surfaces or uniform colors result in the highest model activation densities. Based on this, they developed an attack strategy utilizing images with uniform distributions as energy-latency adversarial inputs, defined as $\mathbf{x}_u \sim \mathcal{N}(\mu,\sigma=\frac{2}{255})$, where $\mu$  is optimized using a grid search. Their attack outperforms optimized sponge examples~\cite{shumailov2021sponge}, while requiring significantly less processing time. {Although this method successfully reduced processing time, addressing a limitation of sponge examples, the generated images are easily detectable due to their characteristic uniform probability distribution.}
Cinà {\it et al.}~\cite{cina2022energy} demonstrated the feasibility of increasing the overall energy consumption of the model without sacrificing accuracy through a poisoning attack strategy. Their approach involves manipulating the optimization objective on a subset of training samples by introducing a regularization term designed to minimize sparsity. Taking this approach further, Wang {\it et al.}~\cite{wang2023energy} integrated this poisoning attack into mobile devices, demonstrating its effectiveness on a real-world testbed. {Despite their attack overcoming the limitations associated with sponge and uniform attacks, the associated increase in energy consumption impairs its stealthiness. In contrast, we have developed a backdoor attack strategy that maintains a lower energy consumption on benign inputs and incorporates triggers that are less perceptually noticeable. 
Backdoor attacks~\cite{liu2018trojaning,liu2020reflection,yu2023backdoor, bhalerao2019luminance} involve the stealthy and deliberate insertion of a malicious behavior into the model that can be activated using a trigger at inference.}
\section{Proposed Energy Backdoor Attack}
\label{sec:pro}
\subsection{Problem Formulation}
In our study, we focus on \ac{dnn} models for image classification, denoted by $\mathcal{M}$. Given a clean training dataset $\mathcal{D}_{cl} = \{(\mathbf{x}_i,y_i)\}_{i=1}^{|\mathcal{D}_{cl}|}$ and a trigger $\delta$, let $\mathcal{C}_{cl}$ be a subset representing a fraction $\alpha = \frac{|\mathcal{C}_{cl}|}{|\mathcal{D}_{cl}|}$ of $\mathcal{D}_{cl}$ , $\mathcal{C}_{po} = \{ (\mathbf{x}_j + \delta,y_j)\}_{j=1}^{|\mathcal{C}_{po}|} | \, (\mathbf{x}_j,y_j) \in \mathcal{C}_{cl}$ represents the trigger subset that is obtained by injecting the trigger $\delta$ into the input samples $\mathbf{x}_j$ of $\mathcal{C}_{cl}$. $\theta_{cl}$ represents the obtained weights when training the model under benign settings on $\mathcal{D}_{cl}$. Our attack aims at optimizing the model weights $\theta_{po}$ to maximize the amount of energy consumed when inferring on poisoned/trigger samples \hbox{$\mathbf{x}_{po} \in X_{po} = \{\mathbf{x}_k\}_{k=1}^{|\mathcal{C}_{po}|} | (\mathbf{x}_k,y_k) \in \mathcal{C}_{po}$}. 
However, to optimize the attack's stealthiness on clean samples \hbox{$\mathbf{x}_{cl} \in X_{cl} = \{\mathbf{x}_k\}_{k=1}^{|\mathcal{D}_{cl}|} | (\mathbf{x}_k,y_k) \in \mathcal{D}_{cl}$}, the poisoned backdoor model should exhibit similar energy consumption levels to those observed when testing the default clean model. Ideally, this should be accomplished while maintaining similar performance in terms of accuracy on both clean and trigger samples. 
The optimization objective for the problem becomes:
\begin{equation}
    \label{optim_obj}
    \begin{split}
    \theta_{po}  = & \argmax_{\theta} \;\;  E\{\mathcal{M} (\theta, X_{po})\} \\
             & - \lambda_{CE} \; [\,\mathcal{L}_{CE}(\mathcal{M}(\theta, X_{cl}),Y)\\
             & + \mathcal{L}_{CE}(\mathcal{M}(\theta, X_{po}),Y)\,]\\
             & - \lambda_{cl} \; E\{\mathcal{M} (\theta,X_{cl})\},
    \end{split}
\end{equation}
where $(\lambda_{CE}, \lambda_{cl}) \in \mathbb{R}^{2+}$ are Lagrangian multipliers,
$E$ represents the model's energy consumption, $\mathcal{L}_{CE}$ refers to the cross-entropy loss and $Y$ is the ground-truth labels.

It is crucial to note that in a sparsity-based accelerator setup, the number of firing neurons within the network for each input is directly proportional to the number of operations performed by the system, and therefore to the energy consumed. Consequently, we quantify energy consumption as the ratio of fired neurons to the total number of neurons. This aligns with the objective formulated in~\cite{shumailov2021sponge,cina2022energy,muller2024impact}, where they optimized an estimate of the hidden activations' $L_0$ norm. The use of a proxy estimation is due to the NP-hard computational intractability of $L_0$-norm optimization. 
Thus, in our experiments, we use the estimation proposed by~\cite{osborne2000lasso} and adopted in~\cite{cina2022energy}, as shown in~\eqref{l0_norm}, to optimize energy.
\begin{equation}
    \label{l0_norm}
    \hat{L}_0(a^{(l)}) = \sum_{i=1}^{K} \frac{(a_i^{(l)})^2}{(a_i^{(l)})^2+\varepsilon}, \; \; \varepsilon \in \mathbb{R^+},
\end{equation}
where $K$ denotes the number of neurons in the $l$-th layer of the model, and $\varepsilon$ is a non-zero constant added to prevent eventual null values at the denominator. 
\begin{figure}[t!]
    \centering
    \includegraphics[width=0.8\linewidth]{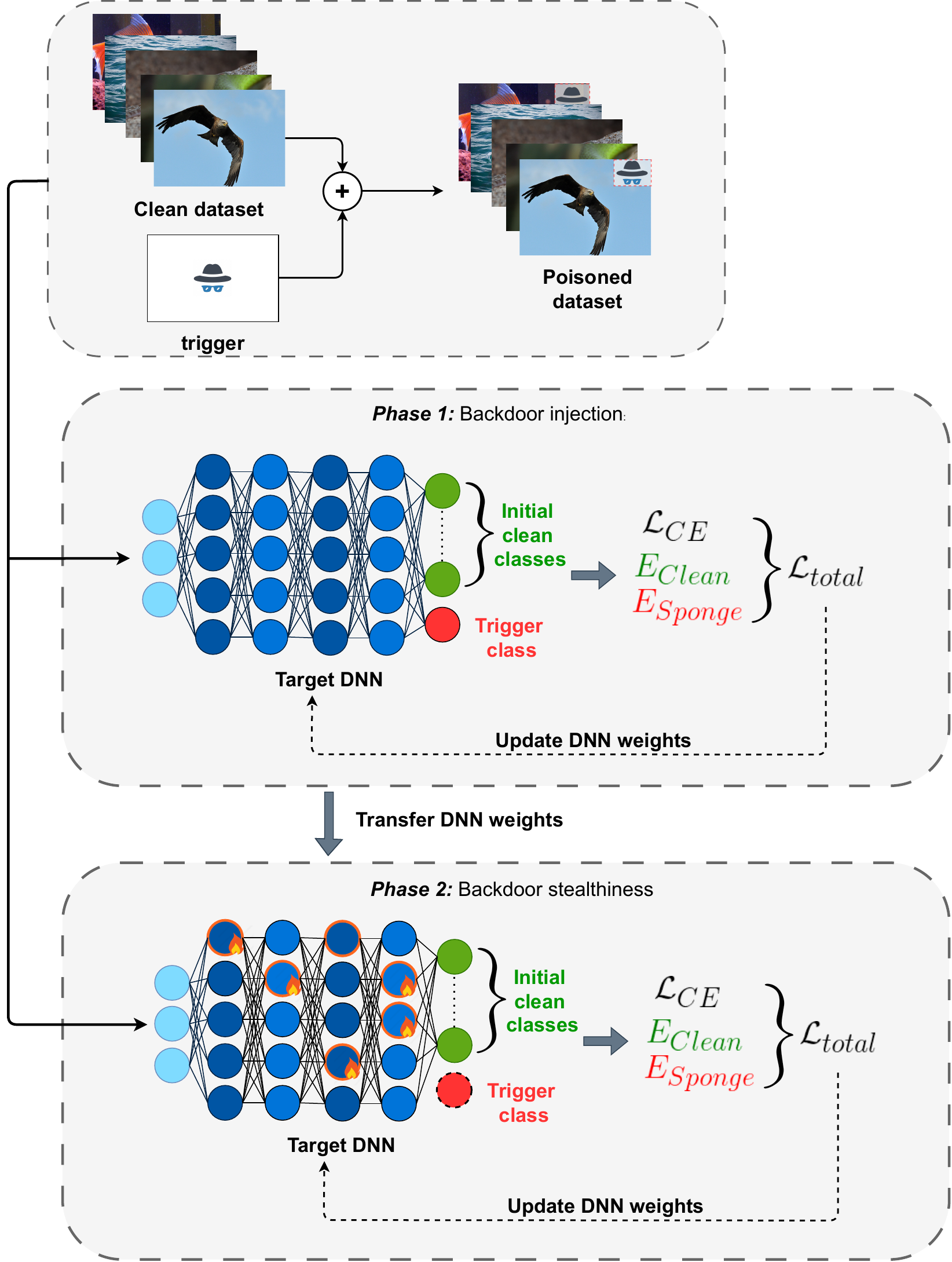}
    \caption{Proposed energy backdoor attack design steps. The adversary first poisons a part of the training set. Then, the backdoor can be injected into the model in two distinct phases. Phase 1 is introduced to ensure effective separability between the clean and trigger samples by adding a supplementary `trigger' class. Next, the `trigger' class is discarded in \hbox{phase 2} and the model obtained in phase 1 is fine-tuned to ensure good accuracy on the trigger samples. 
    }
\label{fig:proposed_approach} 
\vspace{-4mm}
\end{figure}
\subsection{Attack Design}
\label{subsec:attack_design}
Fig.~\ref{fig:proposed_approach} illustrates the training steps of our proposed attack. Given the complexity of solving this particular multi-objective problem, we added a separability objective between the clean and trigger samples' feature maps to guide the model towards a more favorable solution. However, rather than directly embedding this separability objective into the model's loss function, we propose to let the model learn the optimal separability. To achieve this, we split the training process into two phases: \textit{backdoor injection} and \textit{backdoor stealthiness}.
\\
\noindent \textbf{1) Backdoor injection.} The primary goal of this initial stage is to allow the model to learn to differentiate between clean and trigger samples and to learn malicious behavior when fed trigger samples.
First, we introduced an additional class into the training process to specifically annotate the trigger samples.
Subsequently, we modified $C_{po}$ to further incorporate our trigger class $y_{tr}$, i.e., $C_{po}=\{ (\mathbf{x}_j + \delta,y_{tr})\}_{j=1}^{|\mathcal{C}_{po}|} | \, \mathbf{x}_j \in \mathcal{C}_{cl}$.
Then, we proceed to train the model based on the optimization objectives outlined in~\eqref{optim_obj}.
A rudimentary approach to accomplish the latter goal could involve incorporating a similarity minimization term into the backdoor model's training loss to maximize the pairwise distance between clean and trigger feature maps. However, this approach does not guarantee effective separability as the model may push the trigger sample into a feature space of a class that is different from its ground truth label. In contrast, introducing a special class representing the \textit{trigger} samples ensures efficient separation between the extracted features of the trigger and clean samples.
\\
\noindent \textbf{2) Backdoor stealthiness}: The second stage involves fine-tuning the model to enhance the accuracy in classifying the trigger samples and subsequently, improve the stealthiness at inference time. To achieve this, we first excluded the trigger class. This implies replacing the trigger labels in $C_{po}$ by the original ground truth labels. Then, we performed fine-tuning to ensure high accuracy specifically on trigger samples while maintaining the energy optimization objectives in our loss function.
\section{Experimental Results}
\label{sec:exp}

\begin{table*}[t!]
\centering
\caption{Performance comparison of the proposed attack on CIFAR-10 and Tiny-ImageNet datasets. The highest registered energy rates and densities are marked in \textbf{bold}.}
\label{tab:tab}
\resizebox{0.85\textwidth}{!}{%
\begin{tabular}{ccllcccc}
\toprule
Backbone &
 Dataset &{Models}
    &  
   & Accuracy &
  Energy ratio {[}min , max{]} &
  Post-ReLU density {[}min , max{]} &
  Overall density {[}min , max{]} \\\midrule
 &
   &
  \multicolumn{2}{l}{Baseline} &
  94.49\% &
  \multicolumn{1}{c}{86.59\% $\in$ {[}84.96 , 88.30{]}} &
  \multicolumn{1}{c}{0.588 $\in$ {[}0.537 , 0.641{]}} &
  \multicolumn{1}{c}{0.883 $\in$ {[}0.869 , 0.898{]}} \\
 &
   &
  \multicolumn{2}{l}{Sponge-LBFGS~\cite{shumailov2021sponge}} &
  - &
  \multicolumn{1}{c}{86.71\% $\in$ {[}80.96 , 87.11{]}} &
  \multicolumn{1}{c}{0.589 $\in$ {[}0.403 , 0.602{]}} &
  \multicolumn{1}{c}{0.897 $\in$ {[}0.897 , 0.897{]}} \\
 &
 &
  \multicolumn{2}{l}{Sponge-GA~\cite{shumailov2021sponge}}  &
  - &
  87.71\% $\in$ {[}87.70 , 87.72{]} &
  \multicolumn{1}{c}{0.624 $\in$ {[}0.623 , 0.624{]}} &
  \multicolumn{1}{c}{0.893 $\in$ {[}0.893 , 0.893{]}} \\
  &
   &
  \multicolumn{2}{l}{U-inputs~\cite{muller2024impact}} &
  - &
  \multicolumn{1}{c}{88.46\% $\in$ {[}88.45 , \textbf{88.48}{]}} &
  \multicolumn{1}{c}{0.643 $\in$ {[}0.643 , \textbf{0.644}{]}} &
  \multicolumn{1}{c}{\textbf{0.900} $\in$ {[}0.900 , 0.900{]}} \\
 &
   &
   &
  \cellcolor[HTML]{CFE2F3}clean &
  \cellcolor[HTML]{CFE2F3}94.36\% &
  \multicolumn{1}{c}{\cellcolor[HTML]{CFE2F3}86.29\% $\in$ {[}84.58 , 87.94{]}} &
  \multicolumn{1}{c}{\cellcolor[HTML]{CFE2F3}0.578 $\in$ {[}0.527 , 0.630{]}} &
  \multicolumn{1}{c}{\cellcolor[HTML]{CFE2F3}0.881 $\in$ {[}0.866 , 0.895{]}} \\
 &
  \multirow{-6}{*}{{\textbf{CIFAR-10}}} &
  \multirow{-2}{*}{Ours} &
  \cellcolor[HTML]{F4CCCC}trigger &
  \cellcolor[HTML]{F4CCCC}92.64\% &
  \multicolumn{1}{c}{\cellcolor[HTML]{F4CCCC}87.30\% $\in$ {[}86.15 , 88.34{]}} &
  \multicolumn{1}{c}{\cellcolor[HTML]{F4CCCC}0.610 $\in$ {[}0.574 , 0.643{]}} &
  \multicolumn{1}{c}{\cellcolor[HTML]{F4CCCC}0.890 $\in$ {[}0.880 , 0.899{]}} \\
  \cmidrule(r){2-8}
 &
   &
  \multicolumn{2}{l}{Baseline} &
  72.90\% &
  86.80\% $\in$ {[}85.03 , 88.81{]} &
  \multicolumn{1}{c}{0.592 $\in$ {[}0.532 , 0.656{]}} &
  0.885 $\in$ {[}0.870 , 0.903{]} \\
 &
   &
  \multicolumn{2}{l}{Sponge-LBFGS~\cite{shumailov2021sponge}} &
  - &
  82.15\% $\in$ {[}81.67 , 87.96{]} &
  \multicolumn{1}{c}{0.439 $\in$ {[}0.424 , 0.628{]}} &
  0.845 $\in$ {[}0.841 , 0.895{]} \\
 &
   &
  \multicolumn{2}{l}{Sponge-GA~\cite{shumailov2021sponge}} &
  - &
  88.28\% $\in$ {[}88.28 , 88.29{]} &
  \multicolumn{1}{c}{0.641 $\in$ {[}0.641 , 0.642{]}} &
  \multicolumn{1}{c}{0.898 $\in$ [0.898 , 0.898]} \\
 &
   &
  \multicolumn{2}{l}{U-inputs~\cite{muller2024impact}} &
  - &
  89.47\% $\in$ {[}89.46 , 89.48{]} &
  \multicolumn{1}{c}{0.675 $\in$ [0.675 , 0.675]} &
  \multicolumn{1}{c}{0.909 $\in$ [0.909 , 0.909]} \\
 &
   &
   &
  \cellcolor[HTML]{CFE2F3}clean &
  \cellcolor[HTML]{CFE2F3}68.02\% &
  \multicolumn{1}{c}{\cellcolor[HTML]{CFE2F3}87.81\% $\in$ {[}85.79 , 90.45{]}} &
  \multicolumn{1}{c}{\cellcolor[HTML]{CFE2F3} 0.622 $\in$ {[}0.554 , 0.706{]}} &
  \cellcolor[HTML]{CFE2F3}0.894 $\in$ {[}0.876 , 0.917{]} \\
  
\multirow{-12}{*}{{\textbf{ResNet-18}}} &
  \multirow{-6}{*}{{\textbf{Tiny ImageNet}}} &
  \multirow{-2}{*}{Ours} &
  \cellcolor[HTML]{F4CCCC}trigger &
  \cellcolor[HTML]{F4CCCC}66.14\% &
  \multicolumn{1}{c}{\cellcolor[HTML]{F4CCCC}89.00\% $\in$ {[}87.44 , \textbf{90.79}{]}} &
  \multicolumn{1}{c}{ \cellcolor[HTML]{F4CCCC} 0.660 $\in$ {[}0.608 , \textbf{0.716}{]}} &
  \cellcolor[HTML]{F4CCCC}0.904 $\in$ {[}0.891 , \textbf{0.920}{]} \\
  \midrule
 &
   &
  \multicolumn{2}{l}{Baseline} &
  94.35\% &
  \multicolumn{1}{c}{81.42\% $\in$ {[}79.51 , 83.22{]}} &
  \multicolumn{1}{c}{ 0.646 $\in$ {[}0.617 , 0.679{]}} &
  0.827 $\in$ {[}0.809 , 0.844{]} \\
 &
   &
  \multicolumn{2}{l}{Sponge-LBFGS~\cite{shumailov2021sponge}} &
  - &
  \multicolumn{1}{c}{80.94\% $\in$ {[}74.45 , 81.06{]}} &
  \multicolumn{1}{c}{0.638 $\in$ {[}0.522 , 0.640{]}} &
  0.823 $\in$ {[}0.762 , 0.824{]} \\
 &
   &
  \multicolumn{2}{l}{Sponge-GA~\cite{shumailov2021sponge}} &
  - &
  \multicolumn{1}{c}{81.33\% $\in$ {[}81.27 , 81.40{]}} &
  0.657 $\in$ {[}0.655 , 0.657{]} &
  0.826 in {[}0.825 , 0.827{]} \\
 &
   &
  \multicolumn{2}{l}{U-inputs~\cite{muller2024impact}} &
  - &
  80.32\% $\in$ {[}80.28 , 80.35{]} &
  0.640 $\in$ {[}0.639 , 0.640{]} &
  0.817 $\in$ {[}0.816 , 0.817{]} \\
 &
   &
   &
  \cellcolor[HTML]{CFE2F3}clean &
  \cellcolor[HTML]{CFE2F3}93.15\% &
  \cellcolor[HTML]{CFE2F3}81.99\% $\in$ {[}80.10 , 84.33{]} &
  \cellcolor[HTML]{CFE2F3}0.655 $\in$ {[}0.623 , 0.696{]} &
  \cellcolor[HTML]{CFE2F3}0.833 $\in$ {[}0.815 , 0.855{]} \\
 &
  \multirow{-6}{*}{{\textbf{CIFAR-10}}} &
  \multirow{-2}{*}{Ours} &
  \cellcolor[HTML]{F4CCCC}trigger &
  \cellcolor[HTML]{F4CCCC}91.97\% &
  \cellcolor[HTML]{F4CCCC}83.13\% $\in$ {[}81.57 , \textbf{84.77}{]} &
  \cellcolor[HTML]{F4CCCC}0.676 $\in$ {[}0.649 , \textbf{0.707}{]} &
  \cellcolor[HTML]{F4CCCC}0.843 $\in$ {[}0.829 , \textbf{0.859}{]} \\  \cmidrule{2-8}

 &
   &
  \multicolumn{2}{l}{Baseline} &
  72.66\% &
  80.73\% $\in$ {[}78.90 , 81.87{]} &
  0.628 $\in$ {[}0.595 , 0.655{]} &
  0.821 $\in$ {[}0.804 , 0.832{]} \\
 &
   &
  \multicolumn{2}{l}{Sponge-LBFGS~\cite{shumailov2021sponge}} &
  - &
  77.49\% $\in$ {[}74.78 , 80.96{]} &
  0.575 $\in$ {[}0.526 , 0.639{]} &
  0.791 $\in$ {[}0.766 , 0.823{]} \\
 &
   &
  \multicolumn{2}{l}{Sponge-GA~\cite{shumailov2021sponge}} &
  - &
  \multicolumn{1}{c}{80.36\% $\in$ {[}80.35 , 80.38{]}} &
  \multicolumn{1}{c}{0.645 $\in$ {[}0.645 , 0.646{]}} &
  \multicolumn{1}{c}{0.817 $\in$ [0.817 , 0.817]} \\
 &
   &
  \multicolumn{2}{l}{U-inputs~\cite{muller2024impact}} &
  - &
  81.17\% $\in$ {[}81.16 , 81.19{]} &
  \multicolumn{1}{c}{0.656 $\in$ [0.656 , 0.656]} &
  \multicolumn{1}{c}{0.825 $\in$ [0.825 , 0.825]} \\
 &
   &
   &
  \cellcolor[HTML]{CFE2F3}clean &
  \cellcolor[HTML]{CFE2F3}69.39\% &
  \cellcolor[HTML]{CFE2F3}82.31\% $\in$ {[}80.38 , 84.05{]} &
  \cellcolor[HTML]{CFE2F3}0.657 $\in$ {[}0.618 , 0.698{]} &
  \multicolumn{1}{c}{\cellcolor[HTML]{CFE2F3}0.836 $\in$ {[}0.818 , 0.852{]}} \\

\multirow{-12}{*}{{\textbf{MobileNet-V2}}} &
  \multirow{-6}{*}{{\textbf{Tiny ImageNet}}} &
  \multirow{-2}{*}{Ours} &
  \cellcolor[HTML]{F4CCCC}trigger &
  \cellcolor[HTML]{F4CCCC}60.66\% &
  \cellcolor[HTML]{F4CCCC}83.23\% $\in$ {[}81.60 , \textbf{84.50}{]} &
  \cellcolor[HTML]{F4CCCC}0.675 $\in$ {[}0.644 , \textbf{0.701}{]} &
  \cellcolor[HTML]{F4CCCC}0.844 $\in$ {[}0.829 , \textbf{0.856}{]}\\
  \bottomrule
\end{tabular}%
} \vspace{-4mm}
\end{table*}

\subsection{Experimental Setup}
\noindent \textbf{Datasets and baselines.}
Given the distinct nature of our attack that occurs during the training stage and the large size of the ImageNet dataset, we were unable to follow the evaluation setup adopted in~\cite{shumailov2021sponge,muller2024impact}. Instead,
we assessed the performance of our proposed attack on two commonly used datasets for the backdoor attacks evaluation~\cite{zhang2024fiba,li2020invisible,gao2024backdoor,shafahi2018poison,guo2022overview}: CIFAR-10~\cite{cifar10} and Tiny-ImageNet~\cite{tinyImageNet}.
In addition, we employed two deep \ac{cnn} architectures for image classification, namely ResNet-18~\cite{resnet} and MobileNet-V2~\cite{mobilenets}.
We conducted a comparative analysis of the proposed attack against a baseline model to assess its stealthiness, as well as the uniform inputs attack proposed in~\cite{muller2024impact}, and the two configurations of the sponge examples outlined  in~\cite{shumailov2021sponge}, specifically, the \acs{lbfgs} and \acs{ga} setups. The baseline model was trained on the clean dataset $\mathcal{D}_{cl}$ to minimize cross-entropy loss. The uniform inputs and sponge examples with the two configurations referred to as \textit{U-input}, \textit{Sponge-LBFGS} and \textit{Sponge-GA}, respectively, were generated using the provided implementations in~\cite{muller2024impact}. Considering the random nature of these generated inputs, we report the performance obtained on 100 distinct samples, reporting the mean, the maximum and minimum intervals. To ensure a fair comparison, we also selected the optimal $\mu$ parameter for the \textit{U-input} attack using grid search as described in~\cite{muller2024impact}. 
Specifically, we set $\mu=0.2$ for ResNet-18 trained on CIFAR-10, $\mu=0.1$ for ResNet-18 trained on Tiny ImageNet, and $\mu=0$ for MobileNetV2 on both datasets.
In contrast, the reported performance of our proposed attack is derived from computations on the test sets, where `clean' denotes unaltered images, and the `trigger' test set is constructed by adding the trigger to the clean images.

\noindent \textbf{Evaluation metrics.} To evaluate the effectiveness of our attack, we focused on the energy ratio, post-ReLU density, and overall density. This emphasis aligns with previous energy attacks that exploit activation sparsity~\cite{shumailov2021sponge,muller2024impact,cina2022energy}. The energy ratio quantifies the energy costs on an \acs{asic} that incorporates data sparsity optimizations compared to the cost on an \acs{asic} without such optimizations. Additionally, we used top-1 classification accuracy (Test Acc) to assess the stealthiness of our attack.

\noindent\textbf{Implementation details.}
All baseline and backdoor models were trained using the \acs{sgd} optimizer with a learning rate set to $10^{-3}$ for CIFAR-10 and $10^{-2}$ for Tiny ImageNet.  For our backdoor attack, we employed a predefined ramp trigger signal~\cite{barni2019new} with a parameter $\Delta = \frac{60}{255}$. The trigger samples were generated by adding the trigger to the clean image with an overlay weight factor $\gamma = 0.5$. We set $\varepsilon = 10^{-4}$ for the $L_0$ estimation
and limited the poisoning percentage $\alpha$ to less than $5\%$, in line with standard practices in state-of-the-art poisoning and backdoor attacks~\cite{cina2022energy,chen2023dark}.
\subsection{Results and Analysis}
Table~\ref{tab:tab} presents the results of our backdoor attack along with the related work considered. Accuracy scores for U-inputs and sponge examples are omitted from the reported results as these adversarial examples, representing mostly blank or noisy images, lack corresponding ground truth labels.

\noindent \textbf{Backdoor effectiveness.} 
Overall, our attack demonstrates high effectiveness, achieving lower energy rates on clean samples and higher energy rates on trigger samples, on average, across both architectures and datasets. This highlights the vulnerability of \ac{dl} architectures to energy backdoor attacks.
Furthermore, it is noteworthy that the maximum energy ratios, post-ReLU, and overall densities observed on trigger samples significantly exceed those achieved by state-of-the-art attacks, despite their differing natures. A notable exception is the ResNet-18 model on the CIFAR-10 dataset, where our attack's performance metrics closely align with the U-inputs, with a maximum overall density of 0.899 compared to 0.900.


Moreover, the reported energy ratio and densities for the baseline and backdoored models are derived from the test sets. Specifically, for the trigger test set, the input sample's integrity is preserved to maintain the stealthiness of our attack. This explains the significant variance observed in our reported results compared to the U-inputs and Sponge-GA examples. In contrast, for Sponge-LBFGS, the pronounced variance is likely attributed to the L-BFGS algorithm's sensitivity to initial states. This known limitation~\cite{li2016learning} can also explain the algorithm's divergence under certain initializations, which correspond to the minimum energy ratio observed in Table~\ref{tab:tab}.

\begin{figure}[t!]
    \centering
    \includegraphics[width=1\linewidth]{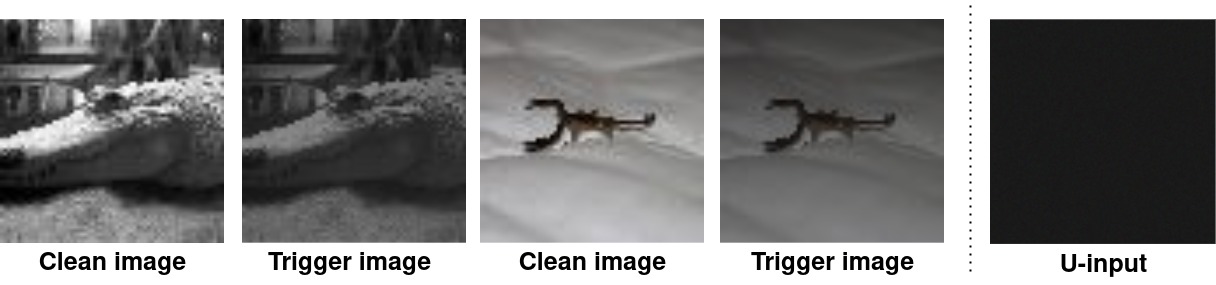}
    \caption{Perceptual comparison between a clean, trigger, and uniform input.}
    \label{fig:samples} \vspace{-5mm}
\end{figure}

\noindent \textbf{Backdoor stealthiness.} To maintain the stealthiness of our attack and avoid raising user suspicion, our backdoored model is designed to meet two essential criteria: 1) exhibit behavior similar to the baseline in terms of both energy consumption and accuracy on clean samples, and 2) maintain high accuracy on trigger examples.
As shown in Table~\ref{tab:tab}, 
we achieved accuracy results remarkably comparable to the baseline, particularly for  backdoored models trained on the CIFAR-10 dataset. Notably, we achieved slightly lower energy consumption than the baseline on clean samples for ResNet-18 trained on CIFAR-10 (86.13\% for ours vs. 86.59\% for the baseline), and comparable energy consumption to the baseline for clean samples using MobileNet-V2. However, we observe lower accuracy rates on backdoored models trained on the Tiny ImageNet dataset, especially on trigger samples. This can be attributed to both the challenging nature of the Tiny ImageNet classification task, which maintains a diverse set of classes despite having fewer categories than the full ImageNet, and the complexity of the objective function relative to the model's capacity. The diversity of the Tiny ImageNet dataset makes it difficult for relatively shallow models to simultaneously learn robust features for effective classification and the desired backdoor behavior. As indicated in~\cite{szegedy2015going}, the number of model parameters is directly related to its capacity to learn high-level and complex representations, which in our scenario would be ideal for handling competing tasks like minimizing energy on clean samples and maximizing it on trigger samples. The model may struggle to achieve optimal performance while adhering to these multiple constraints, as evidenced by a baseline accuracy of only 73\%. Importantly, our proposed attack preserves a natural perceptual aspect of the input, making it less detectable. Fig.~\ref{fig:samples} provides a visual comparison between a clean, trigger, and uniform input. \vspace{-1mm}

\section{Conclusion}
\label{sec:con}
In this paper, we exposed the vulnerability of \acsp{dnn} to {energy backdoor attacks}. We have introduced a novel framework to seamlessly integrate an energy backdoor attack into \acsp{dnn} without compromising the overall accuracy of the model. This model is specifically designed to alleviate the optimization introduced by sparsity-based accelerators when triggered. Despite being confined to the use of a conventional trigger, our experiments confirm the effectiveness and stealthiness of our attack. 
As we anticipate this work will serve as a catalyst for future backdoor attacks focused on energy efficiency, we are aware of the potential for further improvements. This includes exploring optimal triggers and investigating defense techniques to mitigate such attacks.
\bibliographystyle{IEEEbib}
\bibliography{refs}

\begin{thebibliography}{10}

\bibitem{szegedy2015going}
Christian Szegedy, Wei Liu, Yangqing Jia, Pierre Sermanet, Scott Reed, Dragomir Anguelov, Dumitru Erhan, Vincent Vanhoucke, and Andrew Rabinovich,
\newblock ``Going deeper with convolutions,''
\newblock in {\em Proceedings of the IEEE conference on computer vision and pattern recognition}, 2015, pp. 1--9.

\bibitem{azghadi2020hardware}
Mostafa~Rahimi Azghadi, Corey Lammie, Jason~K Eshraghian, Melika Payvand, Elisa Donati, Bernabe Linares-Barranco, and Giacomo Indiveri,
\newblock ``Hardware implementation of deep network accelerators towards healthcare and biomedical applications,''
\newblock {\em IEEE Transactions on Biomedical Circuits and Systems}, vol. 14, no. 6, pp. 1138--1159, 2020.

\bibitem{parashar2017scnn}
Angshuman Parashar, Minsoo Rhu, Anurag Mukkara, Antonio Puglielli, Rangharajan Venkatesan, Brucek Khailany, Joel Emer, Stephen~W Keckler, and William~J Dally,
\newblock ``Scnn: An accelerator for compressed-sparse convolutional neural networks,''
\newblock {\em ACM SIGARCH computer architecture news}, vol. 45, no. 2, pp. 27--40, 2017.

\bibitem{cina2022energy}
Antonio~Emanuele Cin{\`a}, Ambra Demontis, Battista Biggio, Fabio Roli, and Marcello Pelillo,
\newblock ``Energy-latency attacks via sponge poisoning,''
\newblock {\em arXiv preprint arXiv:2203.08147}, 2022.

\bibitem{shumailov2021sponge}
Ilia Shumailov, Yiren Zhao, Daniel Bates, Nicolas Papernot, Robert Mullins, and Ross Anderson,
\newblock ``Sponge examples: Energy-latency attacks on neural networks,''
\newblock in {\em 2021 IEEE European symposium on security and privacy (EuroS\&P)}. IEEE, 2021, pp. 212--231.

\bibitem{chen2023dark}
Simin Chen, Hanlin Chen, Mirazul Haque, Cong Liu, and Wei Yang,
\newblock ``The dark side of dynamic routing neural networks: Towards efficiency backdoor injection,''
\newblock in {\em Proceedings of the IEEE/CVF Conference on Computer Vision and Pattern Recognition}, 2023, pp. 24585--24594.

\bibitem{navaneet2023slowformer}
KL~Navaneet, Soroush~Abbasi Koohpayegani, Essam Sleiman, and Hamed Pirsiavash,
\newblock ``Slowformer: Universal adversarial patch for attack on compute and energy efficiency of inference efficient vision transformers,''
\newblock {\em arXiv preprint arXiv:2310.02544}, 2023.

\bibitem{pan2022gradauto}
J.~Pan, Q.~Zheng, Z.~Fan, H.~Rahmani, Q.~Ke, and J.~Liu,
\newblock ``Gradauto: Energy-oriented attack on dynamic neural networks,''
\newblock in {\em European Conference on Computer Vision}. Springer, 2022, pp. 637--653.

\bibitem{liu2023slowlidar}
H.~Liu, Y.~Wu, Z.~Yu, Y.~Vorobeychik, and N.~Zhang,
\newblock ``Slowlidar: Increasing the latency of lidar-based detection using adversarial examples,''
\newblock in {\em Proceedings of the IEEE/CVF Conference on Computer Vision and Pattern Recognition}, 2023, pp. 5146--5155.

\bibitem{haque2022ereba}
M.~Haque, Y.~Yadlapalli, W.~Yang, and C.~Liu,
\newblock ``Ereba: black-box energy testing of adaptive neural networks,''
\newblock in {\em Proceedings of the 44th International Conference on Software Engineering}, 2022, pp. 835--846.

\bibitem{haque2023antinode}
M.~Haque, S.~Chen, W.~Haque, C.~Liu, and W.~Yang,
\newblock ``Antinode: Evaluating efficiency robustness of neural odes,''
\newblock in {\em Proceedings of the IEEE/CVF International Conference on Computer Vision}, 2023, pp. 1507--1517.

\bibitem{lintelo2024spongenet}
Jona~te Lintelo, Stefanos Koffas, and Stjepan Picek,
\newblock ``The spongenet attack: Sponge weight poisoning of deep neural networks,''
\newblock {\em arXiv preprint arXiv:2402.06357}, 2024.

\bibitem{krithivasan2020sparsity}
Sarada Krithivasan, Sanchari Sen, and Anand Raghunathan,
\newblock ``Sparsity turns adversarial: Energy and latency attacks on deep neural networks,''
\newblock {\em IEEE Transactions on Computer-Aided Design of Integrated Circuits and Systems}, vol. 39, no. 11, pp. 4129--4141, 2020.

\bibitem{huang2024sponge}
B.~Huang, L.~Pang, A.~Fu, S.~Al-Sarawi, D.~Abbott, and Y.~Gao,
\newblock ``Sponge attack against multi-exit networks with data poisoning,''
\newblock {\em IEEE Access}, 2024.

\bibitem{krithivasan2022efficiency}
S.~Krithivasan, S.~Sen, N.~Rathi, K.~Roy, and A.~Raghunathan,
\newblock ``Efficiency attacks on spiking neural networks,''
\newblock in {\em Proceedings of the 59th ACM/IEEE Design Automation Conference}, 2022, pp. 373--378.

\bibitem{chen2022nicgslowdown}
S.~Chen, Z.~Song, M.~Haque, C.~Liu, and W.~Yang,
\newblock ``Nicgslowdown: Evaluating the efficiency robustness of neural image caption generation models,''
\newblock in {\em Proceedings of the IEEE/CVF Conference on Computer Vision and Pattern Recognition}, 2022, pp. 15365--15374.

\bibitem{wang2023energy}
Zijian Wang, Shuo Huang, Yujin Huang, and Helei Cui,
\newblock ``Energy-latency attacks to on-device neural networks via sponge poisoning,''
\newblock in {\em Proceedings of the 2023 Secure and Trustworthy Deep Learning Systems Workshop}, 2023, pp. 1--11.

\bibitem{muller2024impact}
Andreas M{\"u}ller and Erwin Quiring,
\newblock ``The impact of uniform inputs on activation sparsity and energy-latency attacks in computer vision,''
\newblock {\em arXiv preprint arXiv:2403.18587}, 2024.

\bibitem{liu2018trojaning}
Yingqi Liu, Shiqing Ma, Yousra Aafer, Wen-Chuan Lee, Juan Zhai, Weihang Wang, and Xiangyu Zhang,
\newblock ``Trojaning attack on neural networks,''
\newblock 2017.

\bibitem{liu2020reflection}
Yunfei Liu, Xingjun Ma, James Bailey, and Feng Lu,
\newblock ``Reflection backdoor: A natural backdoor attack on deep neural networks,''
\newblock in {\em Computer Vision--ECCV 2020: 16th European Conference, Glasgow, UK, August 23--28, 2020, Proceedings, Part X 16}. Springer, 2020, pp. 182--199.

\bibitem{yu2023backdoor}
Yi~Yu, Yufei Wang, Wenhan Yang, Shijian Lu, Yap-Peng Tan, and Alex~C Kot,
\newblock ``Backdoor attacks against deep image compression via adaptive frequency trigger,''
\newblock in {\em Proceedings of the IEEE/CVF Conference on Computer Vision and Pattern Recognition}, 2023, pp. 12250--12259.

\bibitem{bhalerao2019luminance}
Abhir Bhalerao, Kassem Kallas, Benedetta Tondi, and Mauro Barni,
\newblock ``Luminance-based video backdoor attack against anti-spoofing rebroadcast detection,''
\newblock in {\em 2019 IEEE 21st International Workshop on Multimedia Signal Processing (MMSP)}, 2019, pp. 1--6.

\bibitem{osborne2000lasso}
Michael~R Osborne, Brett Presnell, and Berwin~A Turlach,
\newblock ``On the lasso and its dual,''
\newblock {\em Journal of Computational and Graphical statistics}, vol. 9, no. 2, pp. 319--337, 2000.

\bibitem{zhang2024fiba}
Lu~Zhang and Baolin Zheng,
\newblock ``Fiba: Federated invisible backdoor attack,''
\newblock in {\em ICASSP 2024-2024 IEEE International Conference on Acoustics, Speech and Signal Processing (ICASSP)}. IEEE, 2024, pp. 6870--6874.

\bibitem{li2020invisible}
Shaofeng Li, Minhui Xue, Benjamin Zi~Hao Zhao, Haojin Zhu, and Xinpeng Zhang,
\newblock ``Invisible backdoor attacks on deep neural networks via steganography and regularization,''
\newblock {\em IEEE Transactions on Dependable and Secure Computing}, vol. 18, no. 5, pp. 2088--2105, 2020.

\bibitem{gao2024backdoor}
Yinghua Gao, Yiming Li, Xueluan Gong, Zhifeng Li, Shu-Tao Xia, and Qian Wang,
\newblock ``Backdoor attack with sparse and invisible trigger,''
\newblock {\em IEEE Transactions on Information Forensics and Security}, 2024.

\bibitem{shafahi2018poison}
Ali Shafahi, W~Ronny Huang, Mahyar Najibi, Octavian Suciu, Christoph Studer, Tudor Dumitras, and Tom Goldstein,
\newblock ``Poison frogs! targeted clean-label poisoning attacks on neural networks,''
\newblock {\em Advances in neural information processing systems}, vol. 31, 2018.

\bibitem{guo2022overview}
Wei Guo, Benedetta Tondi, and Mauro Barni,
\newblock ``An overview of backdoor attacks against deep neural networks and possible defences,''
\newblock {\em IEEE Open Journal of Signal Processing}, vol. 3, pp. 261--287, 2022.

\bibitem{cifar10}
Alex Krizhevsky, Geoffrey Hinton, et~al.,
\newblock ``Learning multiple layers of features from tiny images.(2009),'' 2009.

\bibitem{tinyImageNet}
Amirhossein Tavanaei,
\newblock ``Embedded encoder-decoder in convolutional networks towards explainable ai,''
\newblock {\em arXiv preprint arXiv:2007.06712}, 2020.

\bibitem{resnet}
Kaiming He, Xiangyu Zhang, Shaoqing Ren, and Jian Sun,
\newblock ``Deep residual learning for image recognition,''
\newblock in {\em Proceedings of the IEEE conference on computer vision and pattern recognition}, 2016, pp. 770--778.

\bibitem{mobilenets}
Andrew~G Howard,
\newblock ``Mobilenets: Efficient convolutional neural networks for mobile vision applications,''
\newblock {\em arXiv preprint arXiv:1704.04861}, 2017.

\bibitem{barni2019new}
Mauro Barni, Kassem Kallas, and Benedetta Tondi,
\newblock ``A new backdoor attack in cnns by training set corruption without label poisoning,''
\newblock in {\em 2019 IEEE International Conference on Image Processing (ICIP)}. IEEE, 2019, pp. 101--105.

\bibitem{li2016learning}
Ke~Li and Jitendra Malik,
\newblock ``Learning to optimize,''
\newblock {\em arXiv preprint arXiv:1606.01885}, 2016.

\end{thebibliography}

\end{document}